\documentclass[10pt,journal,compsoc]{IEEEtran}\usepackage[]{graphicx}\usepackage[]{color}
\usepackage{xcolor}
\makeatletter
\def\maxwidth{ %
  \ifdim\Gin@nat@width>\linewidth
    \linewidth
  \else
    \Gin@nat@width
  \fi
}
\makeatother
\AtBeginDocument{%
  \providecommand\BibTeX{{%
    \normalfont B\kern-0.5em{\scshape i\kern-0.25em b}\kern-0.8em\TeX}}}

\usepackage{amsmath}

\usepackage{amssymb}

\usepackage{framed}
\makeatletter
 {\par\unskip\endMakeFramed%
 \at@end@of@kframe}
\makeatother

\definecolor{shadecolor}{rgb}{.97, .97, .97}
\definecolor{messagecolor}{rgb}{0, 0, 0}
\definecolor{warningcolor}{rgb}{1, 0, 1}
\definecolor{errorcolor}{rgb}{1, 0, 0}

\usepackage{alltt}

\usepackage{url}

\makeatletter
\let\@authorsaddresses\@empty
\makeatother

\begin{document}

\title{Evaluating Proposed Fairness Models for Face Recognition Algorithms}

\author{John~J.~Howard*,
		Eli~J.~Laird*,
        Yevgeniy~B.~Sirotin*,
        Rebecca~E.~Rubin,
        Jerry~L.~Tipton,
        and~Arun~R.~Vemury 
\IEEEcompsocitemizethanks{\IEEEcompsocthanksitem J. Howard, E. Laird, Y. Sirotin, R. Rubin, and J. Tipton work at the Identity and Data Sciences Lab // Maryland Test Facility in Upper Malboro, Maryland.  \protect\\
\IEEEcompsocthanksitem A. Vemury works at the United States Department of Homeland Security, Science and Technology Directorate in Washington, DC.\protect\\
\IEEEcompsocthanksitem * First authors contributed equally to this research.  Authors listed alphabetically. E-mail correspondence can be sent to info@mdtf.org}}

\markboth{PRE-PRINT COPY}{}

\IEEEtitleabstractindextext{%
\begin{abstract}

The development of face recognition algorithms by academic and commercial organizations is growing rapidly due to the onset of deep learning and the widespread availability of training data.  Though tests of face recognition algorithm performance indicate yearly performance gains, error rates for many of these systems differ based on the demographic composition of the test set.  These ``demographic differentials'' in algorithm performance can contribute to unequal or unfair outcomes for certain groups of people, raising concerns with increased worldwide adoption of face recognition systems.  Consequently, regulatory bodies in both the United States and Europe have proposed new rules requiring audits of biometric systems for ``discriminatory impacts'' (European Union Artificial Intelligence Act) and ``fairness'' (U.S. Federal Trade Commission).  However, no international standard for measuring fairness in biometric systems yet exists.  This paper characterizes two \textit{proposed} measures of face recognition algorithm fairness (fairness measures) from scientists in the U.S. and Europe, using face recognition error rates disaggregated across race and gender from 126 commercial and open source face recognition algorithms.  We find that both proposed methods have mathematical characteristics that make them challenging to interpret when applied to disaggregated face recognition error rates as they are commonly experienced in practice.  To address this, we propose a set of interpretability criteria, termed the Functional Fairness Measure Criteria (FFMC), that outlines a set of properties desirable in a face recognition algorithm fairness measure.  We further develop a new fairness measure, the Gini Aggregation Rate for Biometric Equitability (GARBE), and show how, in conjunction with the Pareto optimization, this measure can be used to select among alternative algorithms based on the accuracy/fairness trade-space.  Finally, to facilitate the development of more robust and interpretable fairness measures in the face recognition domain, we have open-sourced our dataset of machine-readable, demographically disaggregated error rates.  We believe this is currently the largest such dataset of its kind available to the ML fairness community.

\end{abstract}

}

\maketitle

\section{Introduction}

Facial recognition is the process of identifying individuals using the physiological characteristics of their face~\cite{isoiec2382}.  Humans perform such tasks regularly, using dedicated neural pathways that are part of the larger human visual system~\cite{freiwald2016face}.  Moreover, in 1966 it was first demonstrated that a computer program could also execute face recognition tasks~\cite{bledsoe1966model}.  The efficiency and effectiveness of automated face recognition progressed steadily over the next 50 years, but still largely lagged behind trained human performance~\cite{phillips2000feret, phillips2007frvt, white2015perceptual}.  However, in 2014 convolutional neural nets were first applied to the face recognition problem, allowing them to achieve near human performance for the first time~\cite{taigman2014deepface}.  Subsequently, public facing deployments by both private and government organizations increased considerably.  For example, starting in 2015 face recognition became a common method for gaining access to personal electronic devices, such as laptop, tablets, and smart phones~\cite{Apple, Hello}.  In 2016, face recognition was made a part of both Amazon and Microsoft's cloud platforms~\cite{Rekognition, Azure}, allowing businesses and individuals to access face recognition capabilities for minimal fees.  Those web offerings now list dozens of clients, ranging from smart glasses to retail security applications~\cite{awscustomers}.

Government use of this technology has also expanded, primarily in travel and law enforcement applications.  From 2017 onward the United States (U.S.) Customs and Border Protection agency introduced face recognition based border processing for both entry and exit at over 200 locations~\cite{CBP, tvs2018}.  Similar efforts are planned or have already been executed in the European Union (E.U.)~\cite{euparliment}, Australia~\cite{australiaborder}, and Japan~\cite{japanborder}.  In law enforcement, the principal application of face recognition is lead generation for suspect identification, whereby an unknown individual suspected of committing a crime is searched against a gallery of known individuals to ascertain a lead to their potential identity.  A 2016 study noted that approximately one-in-four U.S. law enforcement agencies had access to this kind of capability~\cite{garvie2016perpetual}.  Similar systems have been introduced by police in the United Kingdom~\cite{metuk}, Australia~\cite{nswfacialrecognition}, and at Interpol~\cite{interpol}.

However, there are also long standing reports of face recognition performance varying for people based on their demographic group membership~\cite{phillips2011other, o2012demographic, klare2012face, ngan2015face, buolamwini2018gender,cook2019demographics,krishnapriya2020issues,howard2019effect,grother2019face3,cavazos2020accuracy}.  Of particular concern is the notion that false match rates in face recognition may run higher for certain groups of people, namely African Americans~\cite{krishnapriya2020issues,howard2019effect}.  Higher false match rates could mean that when face recognition is used to identify certain groups of individuals their chances of experiencing a false positive, i.e. incorrect, identification are higher.  This has the potential to be particularly impactful in law enforcement use of face recognition. Indeed, in the summer of 2020, three instances of wrongful arrest involving face recognition identifications by law enforcement were reported in the U.S., all of them involving African American Males~\cite{HillWrong, jerseyFalseMatch, detroitFalseMatch2}.  These concerns prompted the Association for Computing Machinery, the world's largest body of computer scientists, to urge an immediate suspension to private and governmental use of face recognition technologies, citing ``.. clear bias based on ethnic, racial, gender, and other human characteristics..''~\cite{acmfacerecognition}.  Shortly after, in 2021 the Artificial Intelligence Act was introduced in the E.U, which required biometric systems be audited for ``discriminatory impacts''~\cite{EUAIA}. In the same month, the U.S. Federal Trade Commission released new guidance on AI fairness, highlighting that ``It's essential to test your algorithm [for discrimination] on the basis of race, gender, or other protected classes''~\cite{FTC}.

In response, there has been considerable work around how to train and subsequently demonstrate a ``fair'' face recognition algorithm.  To address the latter, two definitions of ``fairness'' in face recognition applications were proposed by scientists seeking to quantify the equitability, or lack thereof, of various face recognition algorithms.  The first, Fairness Discrepancy Rate (FDR), was proposed by scientists from the Idiap Research Institute, a Swiss artificial intelligence laboratory with a long history of contribution to the field of biometrics~\cite{FDR}.  The second, called the Inequity Rate (IR), was proposed by scientists from the U.S. National Institute of Standards and Technology (NIST)~\cite{INEQ1}, a leading scientific body with over 60 years of biometric test and evaluation experience.

To date, neither of these techniques has been extensively utilized in practice or audited using a large corpus of actual face recognition error rates.  Further, there has been relatively little work to understand the utility of these measures for scoring the fairness of deployed algorithms or for selecting among alternative algorithms during procurement.  To address these gaps, we apply these two fairness measures to  error rates disaggregated across race and gender demographic groups from 126 commercial and open source face recognition algorithms and examine their interpretability for scoring or selecting among face recognition algorithms.  We further develop a new fairness measure, which addresses the identified interpretability gaps and demonstrate how this new measure can be used in conjunction with an optimization technique known as Pareto efficiency to rapidly down-select algorithms that could be considered for deployment.

The remainder of this manuscript is organized as follows. Section~\ref{section:Background} discusses the background information related to this study, such as face recognition, face recognition fairness, and broader fairness efforts in the ML fairness community.  Section~\ref{section:Methods} details the methods used in this audit, including the FDR and IR fairness measures (Sections~\ref{subsection:FDR} and~\ref{subsection:Inequity}, respectively), and also a new fairness measure based on the Gini coefficient, the Gini Aggregation Rate for Biometric Equitability, or GARBE (Section~\ref{subsection:GARBE}).  Section~\ref{section:Methods} also highlights a new set of criteria for interpretable fairness measures (Section~\ref{subsection:FFMC}) and an algorithm down-select protocol based on Pareto optimization (Section~\ref{subsection:Pareto}).  Finally Section~\ref{section:Results} presents the findings of this fairness measures audit and Section~\ref{section:Discussion} presents our conclusions and a discussion.

\section{Background}
\label{section:Background}

\subsection{Face Recognition}
\label{subsection:facerecognition}

Face recognition algorithms operate by generating numerical representations of faces, referred to as templates.  Two face templates can then be compared to produce a similarity score $s$ and if $s$ is greater than some discrimination threshold $\tau$ the corresponding faces are declared to be a ``match'' by the algorithm.  This process can be used in both identification tasks, where an unknown probe face is matched to a gallery of faces, and face verification tasks, where a single face is matched to a claimed identity. The \textit{false match rate} and the \textit{false non-match rate} are two error rates used to measure the foundational accuracy of face recognition algorithms. The false match rate (FMR) measures the proportion of face comparisons between different identities, or non-mated face pairs, that result in a match. The false non-match rate (FNMR) measures the proportion of face comparisons of the same identity, or mated face pairs, that do not result in a match.  FMR and FNMR are specific to a given discrimination threshold $\tau$, which is almost universally set so that FMR $<<$ FNMR. In this paper, we discuss the notion of face recognition fairness with respect to the false match and false non-match rates.

\subsection{Software Fairness Generally}
\label{subsection:softwarefairness}

The fairness of software applications in general has garnered much attention in recent years from organizations across a wide swath of disciplines, including computer science, sociology, policy, and others.  This focus on algorithmic fairness has been spurred by cases of disparate outcomes for members of different demographic groups in AI-driven software applications. Cases of differential demographic outcomes have been documented in loan assessment~\cite{saxena2020fairness}, crime prediction~\cite{proPublicaMachineBias, dressel2018accuracy}, and hiring~\cite{raub2018bots}. One of the most notable examples involves the COMPAS system, an algorithm used to predict criminal recidivism.  The outcomes from the COMPAS algorithm are provided to some U.S. courts during bail sentencing. However, a study performed by~\cite{proPublicaMachineBias}, found that the COMPAS algorithm misclassified African-American individuals as `high risk` nearly twice as much as it did White individuals, resulting in a higher likelihood that African-American defendants were jailed for longer sentences. Another prominent example was a ML powered resume screening tool used by Amazon that was discovered to prefer male candidates over female candidates~\cite{reuters_amazon_hiring}.

These patterns have encouraged efforts across industry, government, and academia to emphasize the develop fair software systems~\cite{nasemHumanAITeaming}. For example, a 2016 White House report on algorithmic systems and civil rights noted as part of its future directions the importance of ``... building systems that support fairness and accountability''~\cite{executive2016big}.  In response, new definitions of ``bias'' began to emerge along with mitigation techniques.  A 2017 survey highlighted a list of 5 cumulative unique  definitions of machine learning bias~\cite{AlgoBiasAutonomousSystems}.  By 2019, this number had grown to sixteen~\cite{acm50YearsUnfairnessML} and the most recent survey (as of this writing) in 2021 documented twenty-three distinct forms of bias that may impact ML systems~\cite{acmFairnessMLSurvey}.

Along with the definitions of bias are many competing definitions of algorithm fairness.  Indeed, a 2018 study identified no less than twenty different definitions of algorithmic fairness and noted that many of these definitions were ``mathematically incompatible''~\cite{verma2018fairness}.  Others, have made this observation as well~\cite{inherentTradeOffsFairness, fairPredwDisparateImpact, Berk2018FairnessIC, acm50YearsUnfairnessML}.  Many of these have historic roots in education testing models developed in the 1960s and 70s~\cite{acm50YearsUnfairnessML}.  A summative work on the topic broadly classified fairness measures that predict a target variable ($Y$), using a classifier ($R$), where a sensitive attribute is involved ($A$) into independence ($R \perp A$), separation ($R \perp A \mid Y$), and sufficiency ($Y \perp A \mid R$), although also noting the impossibility of ``..imposing any two of them simultaneously..'', even at this higher level~\cite{barocas-hardt-narayanan}.

\subsection{Face Recognition Fairness Specifically}
\label{subsection:facefairness}

With competing definitions of fairness, it becomes critical to evaluate the merits of each approach in the context of the real-world scenario to which they might be applied.  This targeted approach to measuring fairness also also allows the ability to assess fairness in dynamic scenarios where multiple forms of system failure can impact different users in different ways. In these scenarios, it is often not a single fairness measure one must consider, but a set, where the different failure cases are weighed in terms of their impact~\cite{acmApplicationOfFairnessNotions}.  Face recognition most certainly meets this criteria, as there are numerous ways in which a system can fail, each with different impacts to different users.  In the law enforcement use case in particular, a false positive identification has the resulting harm of possible false arrest and imprisonment for a member of the community.  A false negative identification, whereby a known suspect in a database is missed, carries the harm of a suspect continuing to be at large in a given community.  The favourable outcome in police use of face recognition is therefore a combination of the probability of two distinct error cases, weighted by some social cost of each error case.  The fair outcome is that this favourable outcome occurs equally often across demographic groups.  We also note that the target variable ($Y$ in~\cite{barocas-hardt-narayanan}) in the face recognition application (facial similarity) is often difficult to establish independently of the classifier outcome $R$, ruling out classes of fairness definitions based on separation and sufficiency requirements.

In the absence of other domain specific guidance on fairness, scientists from NIST and the Swiss Idiap Research Institute have proposed two independent measures of fairness with respect to differential error rates. These two methods are known as the Inequity Rate and Fairness Discrepancy Rate, respectively and are discussed in detail in the following sections.

\section{Methods}
\label{section:Methods}

\subsection{Fairness Discrepancy Rate}
\label{subsection:FDR}

Fairness Discrepancy Rate (FDR) was proposed by scientists at the Idiap Research Institute, a Swiss artificial intelligence laboratory with a long history of biometrics research in November of 2020~\cite{pereira2020fairness}.  It was subsequently published in a leading IEEE biometrics journal in August, 2021~\cite{FDR} as the ``.. first figure of merit in this field'' and highlights that it ''consider[s] the FMR and FNMR trade-off in the demographic differential assessment..''.  Essentially, this metric advocates for calculating the max difference in false match rate (FMR) and false non-match rate (FNMR) performance between any two demographic groups $d_{i}$ and $d_{j}$ and a given discrimination threshold $\tau$.  Those differences are then weighed by parameters $\alpha$ and $\beta = 1-\alpha$, which represent the level of concern applied to differences in FMR and FNMR respectively.  The resulting FDR metric is on a scale of 0 to 1, with 1 being ``fair'' and 0 being ``unfair''~\cite{pereira2020fairness}.  The exact equations for calculating FDR are shown in Equations~\ref{eq:FDR_A} -~\ref{eq:FDR}.

\begin{equation}\label{eq:FDR_A}
	A(\tau) = max(| \textit{FMR}_{d_i}(\tau) - \textit{FMR}_{d_j}(\tau) | ) \;\; \forall d_i,d_j \in D
\end{equation}

\begin{equation}\label{eq:FDR_B}
	B(\tau) = max(| \textit{FNMR}_{d_i}(\tau) - \textit{FNMR}_{d_j}(\tau) | ) \;\; \forall d_i,d_j \in D
\end{equation}

\begin{equation}\label{eq:FDR}
	\textit{FDR}(\tau) = 1 - (\alpha A(\tau) + (1- \alpha) B(\tau))
\end{equation}

\subsection{Inequity Rate}
\label{subsection:Inequity}

The Inequity Rate (IR) was proposed by scientists at NIST, a global leader in face recognition performance and testing in March of 2021~\cite{INEQ1}.  Unlike FDR, the IR metric takes ratio differences between min, max FMR and FNMR rates per demographic groups $d_{i}$ and $d_{j}$.  It then raises these differences to weighing factors $\alpha$ and $(1-\alpha)$ and multiplies the results as shown in Equations \ref{eq:A_i} -~\ref{eq:inequity}.

\begin{equation}\label{eq:A_i}
	A(\tau) = \frac{\max_{d_i} \textit{FMR}_{d_i}(\tau)}{\min_{d_j} \textit{FMR}_{d_j}(\tau)} \;\; \forall d_i,d_j \in D
\end{equation}

\begin{equation}\label{eq:B_i}
	B(\tau) = \frac{\max_{d_i} \textit{FNMR}_{d_i}(\tau)}{\min_{d_j} \textit{FNMR}_{d_j}(\tau)} \;\; \forall d_i,d_j \in D
\end{equation}

\begin{equation}\label{eq:inequity}
	\textit{IR} = A(\tau)^\alpha B(\tau)^{1 - \alpha}
\end{equation}

\subsection{Data}
\label{subsection:Data}

Evaluating the properties of summative measures of face recognition fairness requires data.  In the case of the FDR and the IR, the data required must have the properties laid out in Table~\ref{table:DataCriteria}.

\begin{table}[htbp]
\caption{Data criteria for summative face recognition fairness metric evaluation}
\label{table:DataCriteria}
	\begin{center}
	\begin{tabular}{|l|l|}
	\hline
	\textbf{Criteria}			& \textbf{Description} 				\\\hline
	C.1							& False match rates									\\\hline
	C.2							& False non-match rates								\\\hline
	C.3							& Criteria C.1 and C.2 at a single threshold         \\
                                & per algorithm	\\\hline
	C.4							& Criteria C.1 and C.2 dis-aggregated                \\
                                & by demographic group				\\\hline
	C.5							& Criteria C.1 - C.4 across a representative         \\
                                & number of face recognition algorithms \\\hline
	\end{tabular}
	\end{center}
\end{table}

\begin{figure}[t]
  \centering
  \includegraphics[width=\columnwidth]{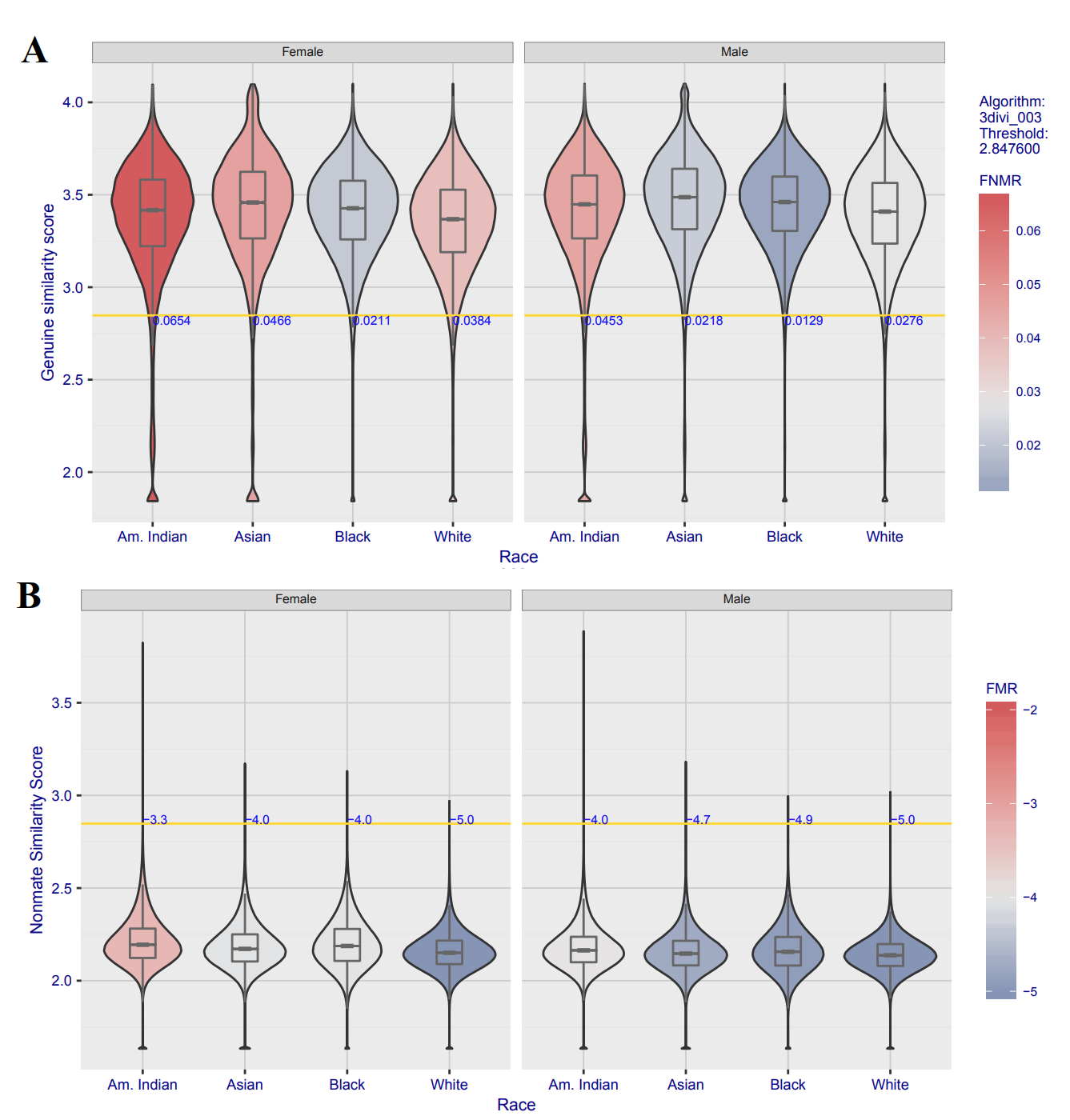}
  \caption{Sample data from NIST FRVT Part 3: Demographics Report, Annex 15}
  \label{figure:FRVT}
\end{figure}

We note this is a non-trivial dataset to develop. Most users and developers of face recognition only have access to a small number of algorithms.  There are also a limited number of large datasets with ground truth demographic data.

However, there is one organization that evaluates a large number of face recognition algorithms and produces such data.  Since 2000, NIST has operated the Face Recognition Vendor Test (FRVT).  The FRVT evaluation is open to face recognition companies and researchers from around the world.  Applicants submit their face recognition algorithm packaged in a NIST defined API.  NIST then runs these algorithms over several large corpora of face images where the identity of the individuals in the photo is known (VISA photos, MUGSHOT photos, WILD photos, etc.).  From these face comparisons, various metrics are produced such as false match and non-match rates at various thresholds.  

Initial FRVT tests were conducted in 2000, 2002, and 2006 as individual test events~\cite{frvt2000, phillips2003face, phillips2007frvt}.  However, since 2018 NIST has run the FRVT evaluation on a ``on-going'' basis, meaning results are released on a rolling schedule several times a year~\cite{grother2018ongoing}.  Newer versions of the FRVT also produce FNMR and FMR metrics disaggregated by demographic factors, such as sex, country of origin, and age, and a fully separate FRVT report focused on demographic factors was released in 2019~\cite{grother2019face3}.  The metrics contained in the FRVT are copious and represent the largest consolidated source of information on the performance of face recognition algorithm.  One section of the report in particular meets the requirements laid out in Table~\ref{table:DataCriteria}.  Annex 15 of~\cite{grother2019face3} contains the data shown in Figure~\ref{figure:FRVT} for each of the 126 algorithms that were analysed in this 2019 report.  They show the mated and nonmated similarity score distribution, as well as the threshold that corresponds to the $1e^{-5}$ FMR for a single demographic.  The other numbers on the horizontal lines are FNMR (plot A) and $log_{10}(\text{FMR})$ (plot B) metrics for eight demographic groups.  The face pairs used to generate these metrics are derived from a subset of a dataset known as "Mugshots", which contains images of individuals involved in routine U.S. law enforcement booking procedures.  Demographic labels are assigned by law enforcement officers and encoded in a record known as the Electronic Biometric Transmission Specification, or EBTS.

For this work, the values contained in NIST FRVT Part 3, Annex 15 were hand transcribed into a machine readable comma separated value file (CSV).  This CSV contains 126 columns (one per algorithm) and 17 rows (algorithm name, 8 false match rates, 8 false non-match rates, one per demographic group).  We believe this is currently the largest, machine readable collection of disaggregated face recognition error rates. We have made this dataset available at our organization's GitHub page for the benefit of the ML fairness community (see Acknowledgements Section).

\subsection{Functional Fairness Measure Criteria}
\label{subsection:FFMC}

One primary objective of any proposed fairness measure is to rank classification algorithms by that measure and select the top or ``most fair''.  We argue this objective is aided when the fairness measure has three properties that make the measure intuitive and more easily reasoned about.  These properties are listed below.  We collectively refer to these three conditions as the Functional Fairness Measure Criteria, or FFMC.

\begin{itemize}
	\item {FFMC.1} - The net contributions of FMR and FNMR differentials to the overall fairness measure should be intuitive when using a normal range of risk parameter weights and operationally relevant error rates.
	\item {FFMC.2} - There should be recognizable points of reference in the domain of the fairness measure.  The easiest way to achieve this objective is to have a bounded fairness measure, with a minimum and maximum possible value.
	\item {FFMC.3} - The fairness measure should be calculable when no errors are observed for a demographic group.  Particularly in the context of face recognition, as an increasing number of intersectional demographic groups are considered, the likelihood of experiencing a group with a FNMR of zero also increases.  Furthermore, in face recognition if cross group FMR numbers are considered, the likelihood of experiencing a group pair with FMR of zero also rises.  The fairness measure should be able to be computed in the presence of either one of these conditions.
\end{itemize}

\subsection{The Gini Aggregation Rate for Biometric Equitability}
\label{subsection:GARBE}

Sections~\ref{subsection:FDRResults} and~\ref{subsection:InequityResults} examine the properties of the FDR and IR metrics using real, disaggregated face recognition error rates against the FFMC criteria.  We find each metric does not fully satisfy the criteria.  We thus propose a third fairness aggregation, called the Gini Aggregation Rate for Biometric Equitability (GARBE), inspired by the mathematics of the Gini coefficient.  The Gini coefficient is a long-standing measure of statistical dispersion of a set of numbers~\cite{gini1912variabilita}.  First applied to income inequality in the 1910s, it has since been utilized by the United Nations~\cite{ungini}, World Bank~\cite{wbgini}, and Organization for Economic Co-operation and Development~\cite{oecdgini} as the premiere measure of wealth disparity.  Moreover, it has been adopted in a number of other fields, including microbiology~\cite{cai2019novel}, cardiology~\cite{sanchez2019introduction}, and ecology~\cite{sun2010application}.  The formula for the generic Gini coefficient, given $n$ observations of a discrete variable $x$ is shown in Equation~\ref{eq:Gini}.  For our purposes, we use a variant that
normalizes the upper bound of the sample by $\frac{n}{n-1}$.  This corrects for downward bias in Gini coefficient calculations when the number of samples is small, as demonstrated in~\cite{deltasGiniCorrection}.

\begin{equation}\label{eq:Gini}
	G_{x} = \left(\frac{n}{n-1}\right) \left(\frac{\sum_{i=1}^{n} \sum_{j=1}^{n} \mid x_{i} - x_{j} \mid } {2n^2\bar{x}}\right) \forall d_i,d_j \in D
\end{equation}

 Given this definition, a simple extension of the Gini coefficient to the face recognition, or general biometric, use case, taking account risk parameters for weighting the impact of a false match versus false non-match error is shown in Equation~\ref{eq:GARBE}.  We coin the term Gini Aggregation Rate for Biometric Equitability (GARBE) to describe this measure.

\begin{equation}\label{eq:GARBE_Terms}
	A(\tau) = G_{\textit{FMR}_{\tau}};\; B(\tau) = G_{\textit{FNMR}_{\tau}}
\end{equation}

\begin{equation}\label{eq:GARBE}
	\textit{GARBE($\tau$)} = \alpha A(\tau) + ( 1-\alpha ) B(\tau)
\end{equation}

\begin{figure}[t]
  \centering
  \includegraphics[width=0.9\columnwidth]{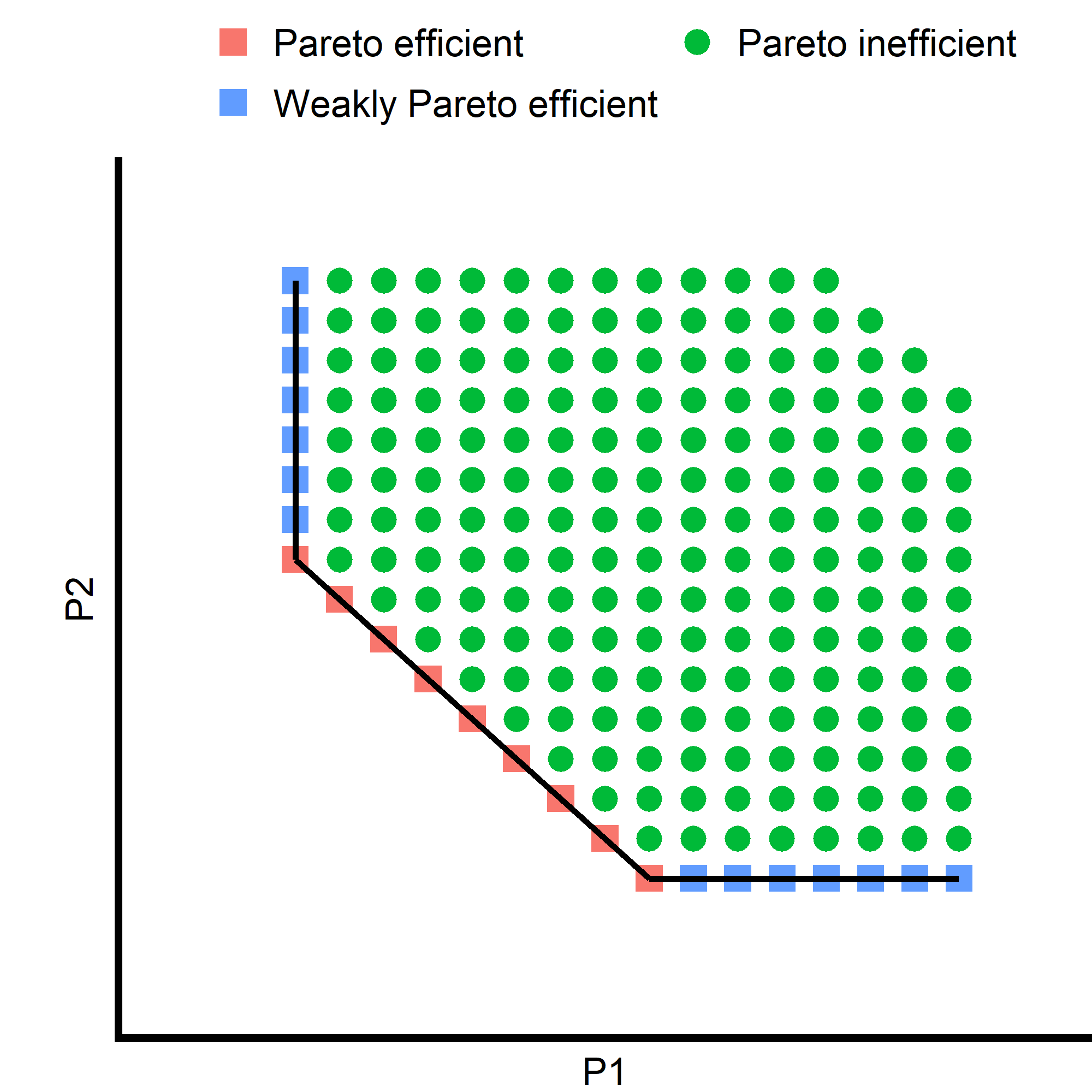}
  \caption{Sample graph showing Pareto efficient and weakly Pareto efficient performance pairs.}
  \label{figure:pareto-sample}
\end{figure}

One potential drawback to the approach proposed by Equations~\ref{eq:Gini}~-~\ref{eq:GARBE} is that various studies have documented grouping effects in Gini calculations that can result in underestimation of numeric dispersion~\cite{ourtiGiniGroupBias, warrensGiniGroupBias}.  For example, consider calculating the Gini coefficient as shown in Equation~\ref{eq:Gini} on error counts as experienced across three groups, A, B, and C.  For the data $x = \{5, 5, 10\}$, the corresponding $G_x = 0.25$.  However, were we to combine the error counts for groups A and B such that $x = \{10, 10\}$ the corresponding would $G_x$ would be 0.  It therefore becomes possible to ``cheat the system'' by grouping the data in such a way that minimizes the Gini coefficient, giving an impression of a ``more fair'' system that would not exist had data been grouped otherwise.  To discourage the intentional use of grouping to bias comparisons involving Gini coefficients, we recommend the specification of grouping variables and group sizes when reporting calculations of the Gini coefficient and derivatives of the metric, such as GARBE. 

\subsection{The Pareto Curve Optimization with Overall Effectiveness}
\label{subsection:Pareto}

As others have noted, fairness is often part of a trade space with another optimization criteria, accuracy~\cite{zafar2017fairness, wei2020fairness}.  For example, one way to achieve ``fairness'' in a face recognition system is to simply declare every face pair as non-matching.  Each demographic group would therefore have precisely equal FNMRs (100\%) and precisely equal FMRs (0\%).  While fair, this solution is less than desirable when one also considers the overall performance of the system.

One common technique for optimization around multiple performance criteria in economics and engineering is Pareto efficiency.  One can say a pair of performance measures for a solution is Pareto efficient if it satisfies the following condition.  Given a set of performance measures $p_{1} = \{ p_{1,1},p_{1,2} ... p_{1,m}\} $ and $p_{2} = \{ p_{2,1},p_{2,2} ... p_{2,m}\} $ for $m$ solutions, a pair $\{p_{1,n},p_{2,n}\}$ is Pareto efficient if both of the following conditions is met:

\begin{equation}\label{eq:Pareto}
    \begin{split}
	p_{1,n} < p_{1,x} \forall x \in \{1,..,m\} \mid x \neq n \\
    p_{2,n} < p_{2,x} \forall x \in \{1,..,m\} \mid x \neq n
    \end{split}
\end{equation}

Similarly, if a pair of performance measures satisfies one condition but not the other then we can say this pair is weakly Pareto efficient (see Figure~\ref{figure:pareto-sample}).

\section{Results}
\label{section:Results}

\subsection{Properties of Fairness Discrepancy Rate in Practice}
\label{subsection:FDRResults}

\begin{figure*}[ht]
  \centering
  \includegraphics[width=0.7\linewidth]{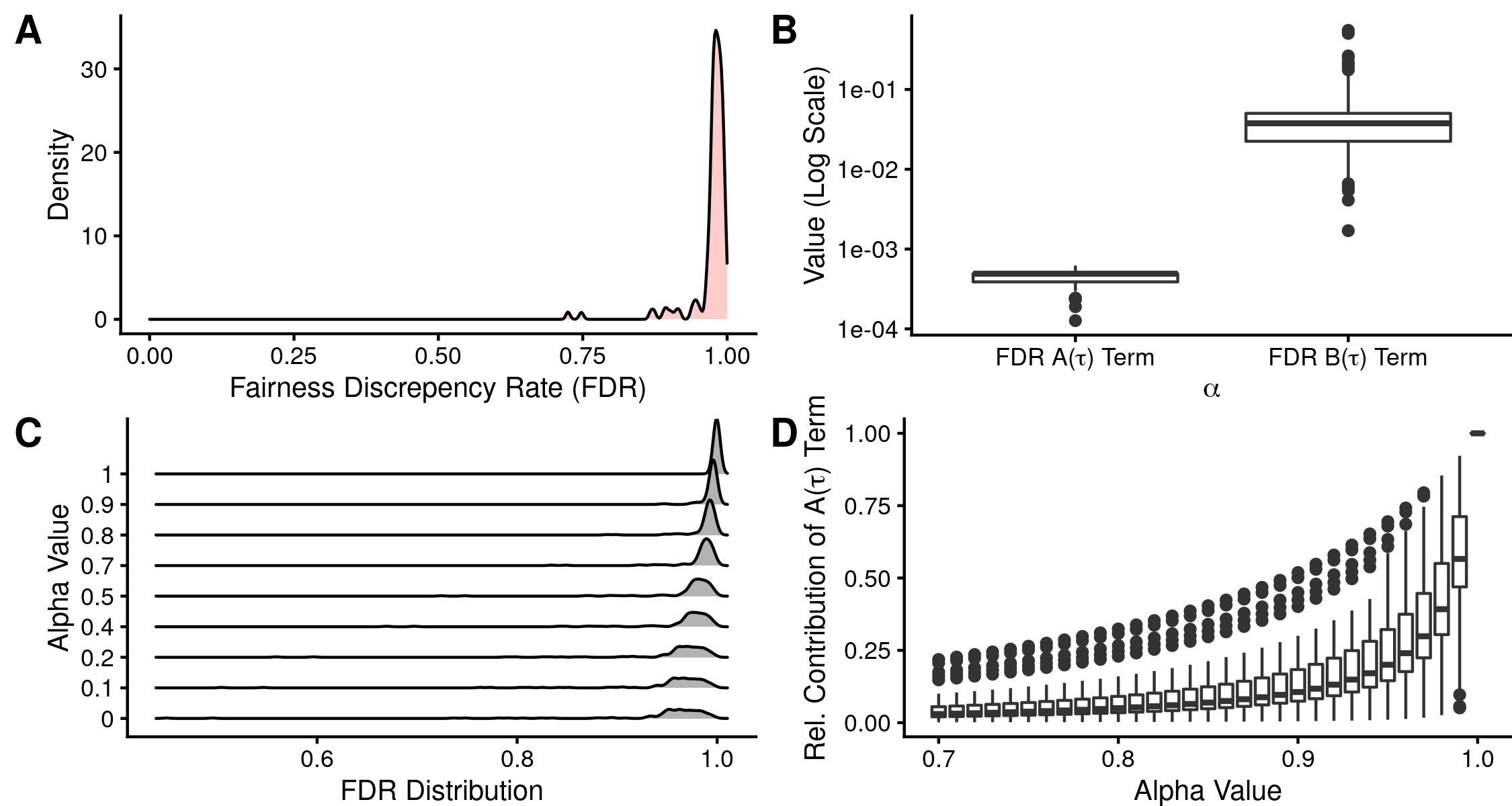}
  \caption{Fairness Discrepancy Rate values using NIST FRVT Part 3 face recognition error rates. \textbf{A.} Overall distribution of FDR values (alpha = 0.5).  \textbf{B} Magnitude of the alpha and beta terms in Equation~\ref{eq:FDR}. \textbf{C.} Minimum and maxiumum values for FDR given an alpha setting. Note the convergence of the range as alpha increases.  \textbf{D.} Relative contribution of the alpha term to the overall FDR value.  Note the truncated x scale (0.7 - 1.0) and that the median contribution of the alpha term does not surpass 50\% until alpha is set to 0.99.}
  \label{figure:FDR}

\end{figure*}

When we apply the data described in Section~\ref{subsection:Data} to the FDR measure (Section~\ref{subsection:FDR}) we see the distribution of FDR measures as shown in Figure~\ref{figure:FRVT}.  We notice that, despite having a theoretical range of 0 to 1, the practical range of the FDR measure, with the alpha and beta set to 0.5, is closer to 0.9 to 1, with over 95\% of FDR values falling in that range.  This is a straightforward mathematical extension of the fact that, while the act of aggregating error rates makes sense in principle, for the face recognition problem in particular these error rates almost always exist on vastly different scales.  For example, using sample data in Figure~\ref{figure:FRVT} we see FNMRs ranging from 1.29\% to 6.54\%.  Conversely, the false non-match rates are orders of magnitude smaller, ranging from 0.001\% ($1e^{-5}$) to approximately 0.05\% ($10^{3.3} ~= 0.000501$).  This is generally true of all face recognition error rates found in our dataset (see Figure~\ref{figure:FDR}B, note the log scale of the y axis).  This has the effect of limiting the FDR measure, for all practical purposes, to 1 minus the difference in FNMR \textit{only}, hence the practical range from 0.9 to 1.0 (FNMR differences typically vary by $<$1\% to 10\%).

Furthermore, this aggregation of error rates that exist on significantly different scales has the extended effect of making the risk parameter $\alpha$ a challenge to configure correctly.  Recall from Section~\ref{subsection:FDR} that alpha is the ``weight'' of the false match discrepancy in the overall FDR calculation.  However, because of the small magnitude of FMR differences, these differences only begin to impact the FDR calculation on an equal scale as FNMR differences when alpha is set to greater than 0.99.  Indeed, from Figure~\ref{figure:FDR}D we see that the median relative contribution of the FMR difference to the FDR only surpasses 50\% when alpha is 0.99 and higher.

\subsection{Properties of Inequity Rate in Practice}
\label{subsection:InequityResults}

Because of the ratio rather than aggregation based summative nature of the Inequity Rate (IR) metric, the issues discussed in Section~\ref{subsection:FDRResults} are largely absent.  The distribution of IR values at the default alpha of 0.5 spans a range from 2.4 to 26.38 with lower values representing is more ``fair'' algorithms in this metric system (Figure~\ref{figure:INEQ}A).  The $A(\tau)$ and $B(\tau)$ terms are on more similar scales, with $A(\tau)$ typically having a value in the 40 - 50 range and the $B(\tau)$ term typically ranging from 4 to 9.  This more congruous relationship between the $A(\tau)$ and $B(\tau)$ terms means the IR reacts to changes in false match rate weight ($\alpha$) with IR distributions continuing to span representative portions of the metric space at all values of $\alpha$ (Figure~\ref{figure:INEQ}C) and the $A(\tau)$ term having more of an impact as alpha rises (Figure~\ref{figure:INEQ}D).

The only challenge to interpreting IR values that arises from this analysis is the unbounded nature of the metric.  Because of its multiplicative nature and the exponential risk weights, there is no theoretical upper bound on the IR measure.  Although the practical upper limit in this study was 63.1, different face recognition algorithms could, in theory, give IR results that approach infinity.  Similarly, this ratio property also has the drawback of making IR incalculable when the min FNMR or FMR for any group is 0.

\subsection{Properties of the Gini Aggregation for Biometric Equitability in Practice}
\label{subsection:GiniResults}

\begin{figure*}[ht]
  \centering
  \includegraphics[width=0.7\linewidth]{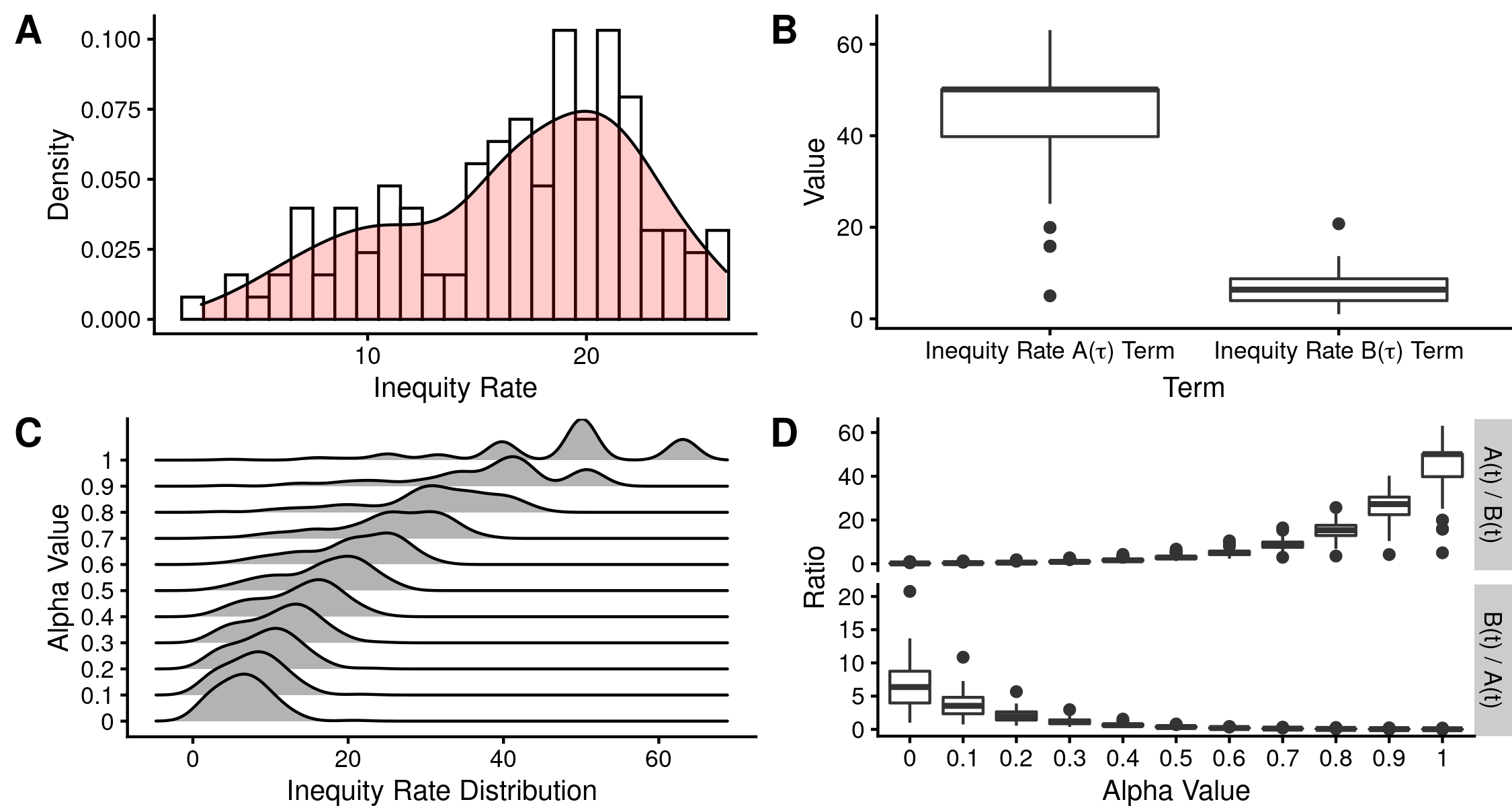}
  \caption{Inequity Rate (IR) values using NIST FRVT Part 3 face recognition error rates.  \textbf{A.} Overall distribution of IR values (alpha = 0.5).  \textbf{B} Magnitude of the alpha and beta terms in Equation~\ref{eq:inequity}. \textbf{C.} Distribution of IR values given an alpha setting.  \textbf{D.} Relative contribution of the alpha term to the overall IR value.}
  \label{figure:INEQ}
\end{figure*}

The Gini Aggregation for Biometric Equitability (GARBE) measure combines the positive characteristics of the FDR and IR measures.  Namely, it's a summative aggregation, meaning the bound can be reasonably controlled but it does not add or subtract error rate values that, in practice, exist on markedly different scales.  Instead the GARBE calculates the Gini coeffecient as an approximation to the ``spread'' or dispersion of these error rates and leverages the fact that the resulting coefficient is already scaled from 0 to 1.  This coefficient can then be weighed using the same basic, multiplicative weighing technique utilized in the FDR metric.  We see that using a default $\alpha$ of 0.5, GARBE metrics for algorithms in~\cite{grother2019face3} span about half of the theoretically usable range (0.165 - 0.618, Figure~\ref{figure:GARBE}A).  This range continues to span representative portions of the metric space as false match error weight ($\alpha$) is modulated (Figure~\ref{figure:GARBE}C).  We also note that the $A(\tau)$ and $B(\tau)$ terms are the only terms in any of the summative fairness measures presented here that are scaled to the same order of magnitude, with the median $A(\tau)$ value found at 0.74 and the median $B(\tau)$ at 0.33 (Figure~\ref{figure:GARBE}B).  Finally, because of the consistent scaling of the Gini coefficient calculation, the relative contribution of the $A(\tau)$ term to the overall  GARBE metric increases approximately linearly as alpha increases (Figure~\ref{figure:INEQ}D), with the mean contribution of  $A(\tau)$ surpassing the contribution of $B(\tau)$ when $\alpha = 0.4$.  Contrast this with Figure~\ref{figure:FDR}D where the mean contribution of $A(\tau)$ did not surpass 0.5 until $\alpha = 0.99$.

\subsection{In Summary of Summative Fairness Measures}
\label{subsection:Summary}

Table~\ref{table:Summary} summarizes the three summative fairness measures investigated by this study against the FFMC interpretability requirements laid out in Section~\ref{subsection:FFMC}.  Because the FDR metric is bounded, we find it is possible to create reference points in its domain.  For example, a perfectly fair algorithm (no differences in group based FNMR or FMR) has a FDR of 1 and an perfectly unfair algorithm (all FNMR or FMR occurring for one group) has a FDR of 0.  FDR is also calculable in the presence of zero percent FNMR or FMR.  However, the FDR measure's differential terms exist at vastly different scales when using a normal range of risk parameters and operationally relevant error rates (Figure~\ref{figure:FDR}B).  In face recognition deployments where false match rate differences across group are of concern, the FDR $alpha$ term should be set on the scale from $(0.99,1]$ in order to allow the contributions from the $A(\tau)$ term to contribute to the overall FDR measure.  This is not documented anywhere outside this audit but is an important point should the FDR measure be used to select fair face recognition algorithms in practice.

The IR fairness measure largely rectifies the scaling issues encountered with FDR measure by taking a ratio as opposed to minmax aggregation of FMR and FNMR numbers.  This results in the IR measure having a dynamic range that spans from a supposed minimum of 1 (``fair'' algorithm) to a practical maximum of 63.1, in this study.  Furthermore, the contribution from $A(\tau)$ and $B(\tau)$ are on relatively similar scales when alpha values are set to normal ranges.  However, also because of this ratio aggregation, the IR measure can approach $\infty$ as $\min_{d_j} \text{FNMR}_{d_j}(\tau)$ or $\min_{d_j} \text{FMR}_{d_j}(\tau)$ approaches 0 and is indeed incalculable should one of these rates reach 0.  Its also challenging to interpret and compare IR values both within and across studies.  Because of the unbounded nature of the measure, the most direct approach to establishing a ``fair'' algorithm is to partition the IR space and select algorithms in the Nth quartile.  However, this quartile can shift from study to study, depending on the minimum FNMR and FMR's per group encountered.  This makes comparing IR values a challenge should the IR measure be used to select fair face recognition algorithm in practice.

\begin{figure*}[ht]
  \centering
  \includegraphics[width=0.7\linewidth]{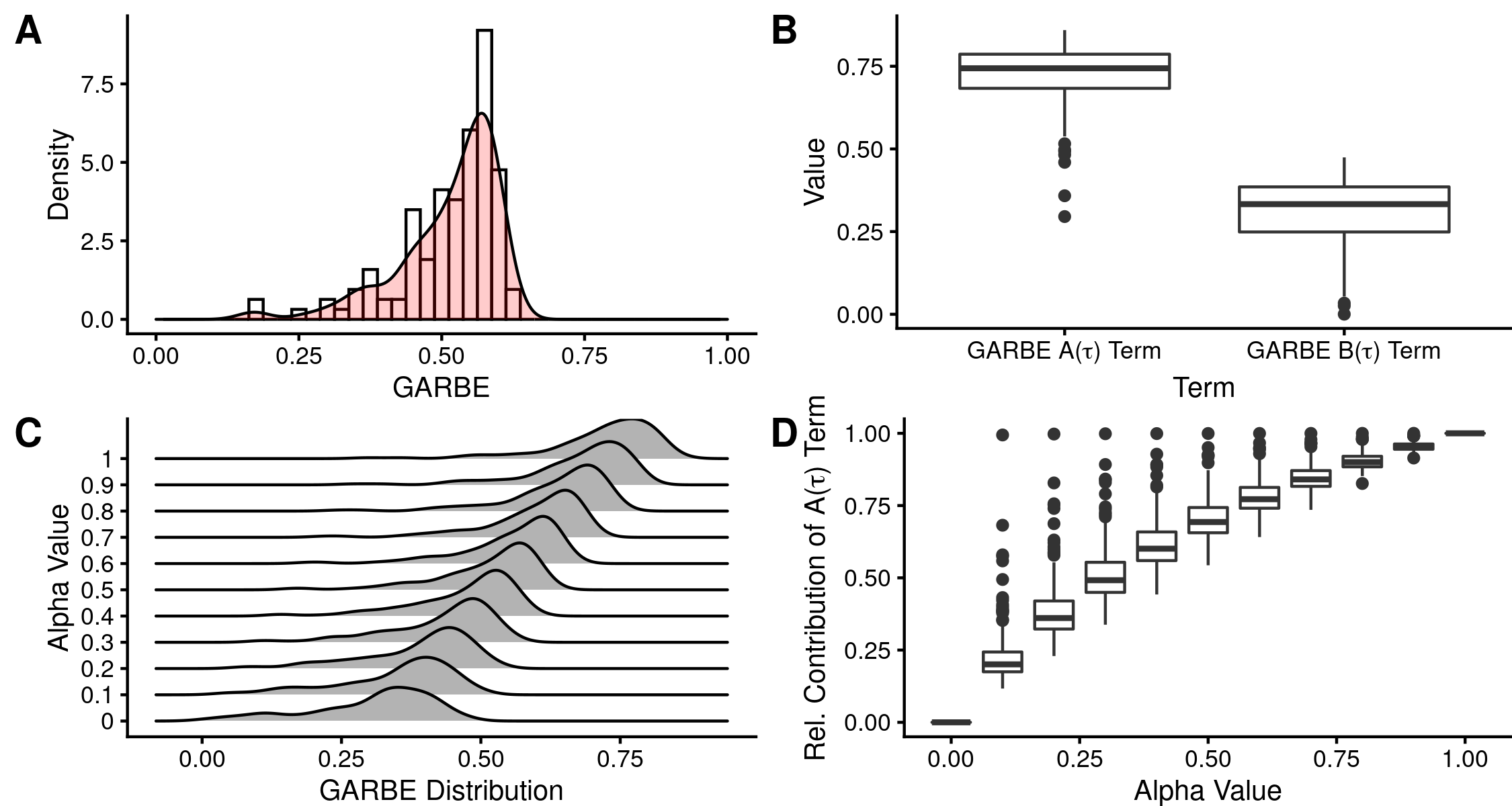}
  \caption{GARBE values using NIST FRVT Part 3 face recognition error rates.  \textbf{A.} Overall distribution of GARBE values (alpha = 0.5).  \textbf{B} Magnitude of the alpha and beta terms in Equation~\ref{eq:GARBE}. \textbf{C.} Distribution of GARBE values given an alpha setting. \textbf{D.} Relative contribution of the alpha term to the overall GARBE value.}
  \label{figure:GARBE}
\end{figure*}

Finally, the GARBE fairness measure, proposed in this study, builds on the strength of the FDR and IR measures.  Instead of aggregating minmax FNMR and FMR differences, the GARBE measure weighs and aggregates measures of dispersion of these error rates, namely the Gini coefficient (Equation~\ref{eq:Gini}).  This has several advantages.  One, this measure is calculable in the presence of error rates being 0.  Second, this measure first converts two sets of numbers that exist on markedly different scales to a single common metric space before weighing and aggregating.  In this fashion we can both avoid the poor relationship between risk ratios and relative contribution of $A(\tau)$ and $B(\tau)$ terms (Figure~\ref{figure:FDR}C-D and~\ref{figure:GARBE}C-D) and retain a bounded domain (Figure~\ref{figure:INEQ}A \&C and~\ref{figure:GARBE}A \& C).  Because of these properties, the GARBE measure is able to satisfy all the FFMC criteria.  Table~\ref{table:Summary} summarizes the three fairness measures with respect to the FFMC criteria.

\begin{table}[htbp]
\caption{Summary of Summative Fairness Measures}
\label{table:Summary}
	\begin{center}
	\begin{tabular}{|l|l|l|l|}
	\hline
	\textbf{FFMC Criteria}	    & \textbf{FDR}		&\textbf{IR}		&\textbf{GARBE}	\\\hline\hline
	FFMC.1					    & 					& \checkmark		& \checkmark				\\\hline
	FFMC.2						& \checkmark     	&                	& \checkmark				\\\hline
	FFMC.3						& \checkmark		& 					& \checkmark				\\\hline
	\end{tabular}
	\end{center}

\end{table}

\subsection{Pareto Curve Optimization with the Gini Aggregation Rate for Biometric Equitability}

Finally, this study advocates for evaluating face recognition algorithms along multiple axes of performance, namely overall effectiveness and fairness, using the Pareto curve method (Section~\ref{subsection:Pareto}).  This has been proposed in other areas of ML fairness more broadly but this is the first study to specifically promote this technique and demonstrate its utility in the context of face recognition tasks~\cite{wei2020fairness, martinez2020minimax, valdivia2021fair}.

This technique requires computing both overall effectiveness and a fairness measure.  We select the GARBE fairness measure for the reasons outlined in Section~\ref{subsection:Summary}.  As a measure of overall performance we select total FNMR across all demographic groups.  This value is a weighted average of the error rates as reported in Annex 15 (Section~\ref{subsection:Data}) and the mated comparison counts, also provided in the introductory material to Annex 15 of~\cite{grother2019face3}.

\begin{figure*}[ht]
  \centering
  \includegraphics[width=0.7\linewidth]{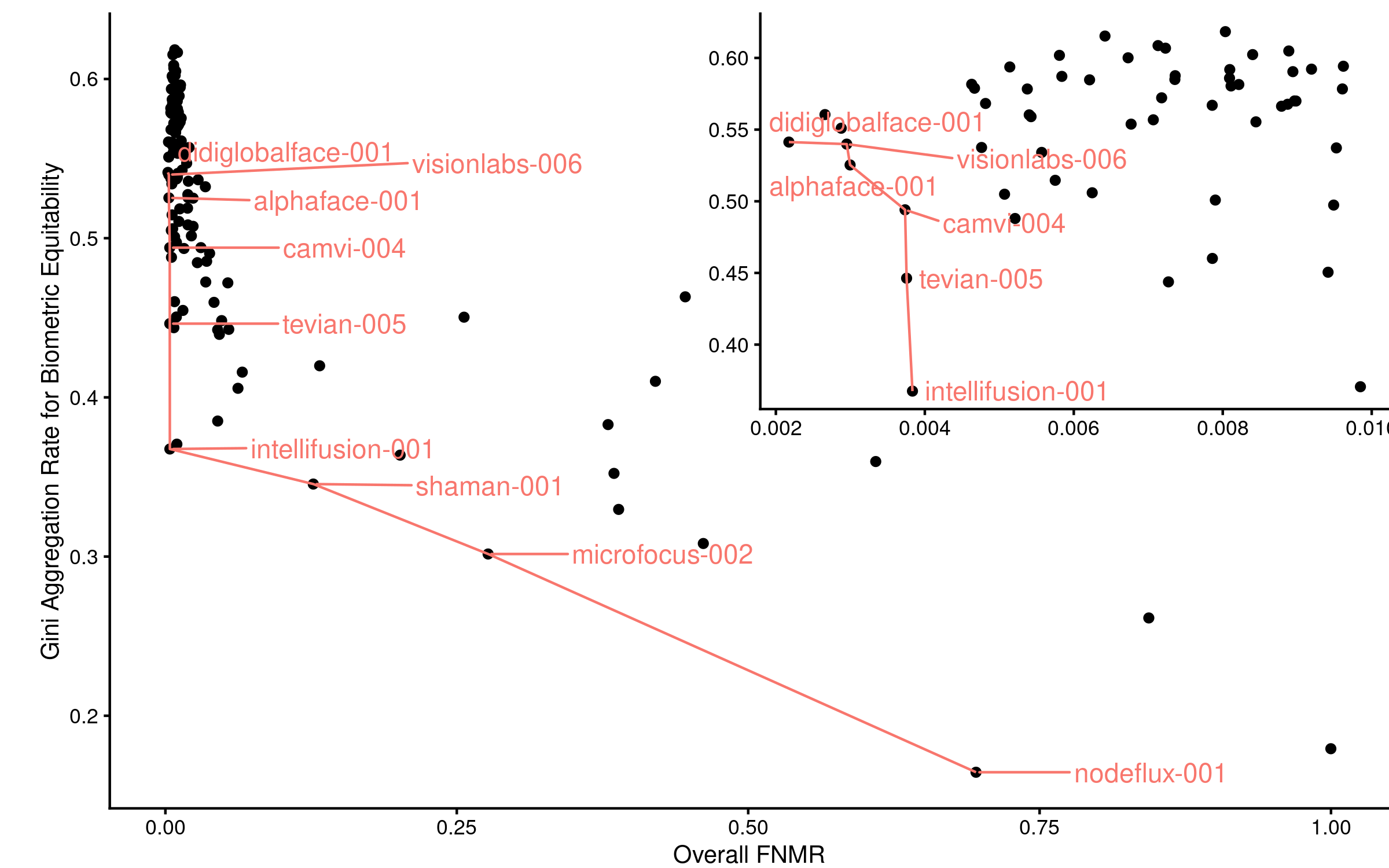}
  \caption{Pareto curve of Gini Aggregation Rate for Biometric Equitability (GARBE) values plotted against overall performance (Total FNMR) using NIST FRVT Part 3 face recognition error rates.  Red line connects algorithms that are Pareto efficient.  No algorithms are weakly Pareto efficient. Inset shows zoom1ed area where total FNMR performance is less than 1\%}
  \label{figure:Pareto}
\end{figure*}

This result is shown in Figure~\ref{figure:Pareto}, with overall performance (FNMR) plotted on the x axis and the GARBE fairness measure plotted on the y axis.  Each point represents an algorithm, while the Pareto efficient algorithms are connected with a red line and have their names printed.  We note the Pareto frontier provides a perceptive means of down-selecting which algorithms should be considered in this optimization space.  Any algorithm not on the Pareto frontier can be discarded, as there exists another selection that is either mathematically more fair or better performing.  This effectively reduces the search space for the ``optimal'' algorithm from the 126 algorithms tested in~\cite{grother2019face3} to the 9 on the Pareto frontier, a savings of over 90\%.  Additionally, if we further refine our search to algorithms that had very good performance overall, we only have to consider the six algorithms in the inset of Figure~\ref{figure:Pareto}.  Algorithm didiglobalface-001 is the highest performing in this space, having achieved the lowest overall FNMR of ${\sim}0.0022$.  However, it is also the least fair of the Pareto efficient set, having achieved the highest GARBE measure of ${\sim}0.54$.  Conversely, algorithm intellifusion-001 was the least performative of this set, with a total FNMR of ${\sim}0.0038$, but it also had a somewhat improved GARBE fairness measure at ${\sim}0.37$.  Whether this trade-off of a 0.0016 increase in total performance is worth a decline in fairness of 0.17 is a question that can be posed to system designers.  However, the Pareto curve, frontier, and process we have outlined here allow this trade-space to be explored effectively.

\section{Discussion}
\label{section:Discussion}

In this study we have executed the first audit of two proposed face recognition fairness measures using demographically disaggregated false match and false non-match error rates from 126 commercial and open source algorithms.  We've found that both proposed models have benefits and drawbacks when it comes to interpreting their outcomes on face recognition error rates commonly found in practice.  We've attempted to consolidate the benefits of each approach into a set of interpretability criteria, called the FFMC, and hope these can serve as a guide for future development of fairness measures, particularly in the face recognition domain.  We've also proposed an alternative fairness measure, the Gini Aggregation Rate for Biometric Equality or GARBE that satisfies all of these criteria and demonstrated a protocol using Pareto efficiency that can rapidly identify optimal algorithms in both the overall performance and fairness domains. The main takeaways and areas of future work are delineated below.

\subsection{Audit the Audit}
\label{subsection:evaluate}

As discussed in Section~\ref{subsection:softwarefairness} and~\ref{subsection:facefairness}, there are currently a plethora of definitions for both bias sources and fairness measures propagating throughout the ML fairness space.  This increased attention is a positive development.  However, in such an environment, it is critical to evaluate the merits of different approaches when applying a given technique to a specific use case.  Often times, when analysing the positive outcome in a specific application of a ML decision system, there is not one specific failure case that can cause harm but a set, which requires the aggregation of different error probabilities into a new metric, as we have shown here.  As new metrics are developed and proposed, purveyors and evaluators of ML algorithms should strive to ensure the statistical properties of their proposed methods are well documented via the kind of audit we have performed here.  In this way they can be of maximum utility to the broader ML fairness community.

\subsection{On the Need for Additional Fairness Data}

One obvious yet often illusive requirement for auditing fairness measures in any domain is data.  This study documented a set of criteria necessary for the evaluation of face recognition fairness measures in particular (Table~\ref{table:DataCriteria}).  Access to data of this nature is a foundational requirement for auditing fairness measures, yet datasets of this nature are limited at best.  We have attempted to provide one such dataset by open sourcing the hand transcribed error rates, per demographic group, used in this research.  However, even this dataset has certain drawbacks.  For example, our dataset only shows error rates at a single population-wide FMR threshold.  FNMR and FMR measures across a range of representative thresholds would allow for a more complete investigation.  Additionally, our dataset only includes intra-demographic false match error rates (Male-to-Male and Female-to-Female, for example).  However, as suggested elsewhere, equatability in the face recognition domain may depend on \textit{inter}-demographic false match rates (Male-to-Female, for example)~\cite{dhs2021quantifying}.  To promote future work in this area, evaluators of face recognition algorithms should consider making more robust datasets available to the community in a readily parsable format.

\subsection{Limitations of Mathematical Formulations of Fairness}

Finally, we conclude with a discussion of the general term ``fairness'' in relation to the kind of mathematical audits we have performed here.  As others have noted, fairness is a broad concept without a concise definition~\cite{verma2018fairness, barocas-hardt-narayanan}.  Additionally, as observed by individuals, fairness is not primarily a mathematical construct but a social and perceptual one~\cite{jacobs2021measurement, kahneman1986fairness}. We've used the term ``fairness measure'' as have others in the sense that these metrics relate to the \textit{topic} of fairness, as they are used to reason about differential error rates.  However, one area that is currently under-researched in the ML fairness community is how mathematical notions of fairness translate to perceptual notions of fairness.  Human perception is often nonlinear and we have accounted for these non-linearities in measurements of physical intensity (e.g. light and sound~\cite{raub2018bots}) and in economic models~\cite{kahneman2013prospect}.  Furthermore, if a system has precisely equal odds that a privileged and unprivileged group will receive a positive outcome in a given fairness space (e.g. a disparate impact of 1), does a human observing this system operate perceive it to be fair?  There very well may be entire classes of AI systems, face recognition included, that regardless of their performance may be perceived as unfair in some applications.  Should this be the case, then, despite current consensus in the literature, the term ``fairness'' may not be appropriate for describing the class of metrics that deal more narrowly with differential performance of the system rather than the perceptual fairness of a particular application of the system.  We think studies to understand human perception of fairness will help bridge current gaps between notions of mathematical fairness based on accuracy and social/perceptual fairness in the ML fairness community.

\section*{Acknowledgments}
\label{section:Acknowledgements}
 This research was sponsored by the Department of Homeland Security, Science and Technology Directorate on contract number W911NF-13-D-0006-0003. The views presented here are those of the authors and do not represent those of the Department of Homeland Security, the U.S. Government, or their employers.  The error rate data used in this research are publicly available from the U.S. National Institute of Standards and Technology in PDF form.  We have parsed these error rates into a machine readable CSV file as part of this research and made this dataset available for the benefit of the ML fairness community at our organization's GitHub page (\url{http://github.mdtf.org}).  The authors are thankful to their colleagues at the NIST Information Technology Laboratory and at the International Organization for Standardization (ISO), who's discussions and thoughts contributed to this work.

\bibliographystyle{IEEEtran}
\bibliography{fairness-manuscript}

\vfill

\end{document}